\documentclass[sn-mathphys,Numbered]{sn-jnl}

\usepackage{graphicx}%
\usepackage{amsmath,amssymb,amsfonts}%
\usepackage{xcolor}%
\usepackage{manyfoot}%
\usepackage{booktabs}%
\usepackage{tabularx}%
\usepackage{enumitem}%
\usepackage{tikz}%

\definecolor{RPDarkRed}{HTML}{BD0519}
\definecolor{RPCrimson}{HTML}{EB301F}
\definecolor{RPOrange}{HTML}{FB8520}
\definecolor{RPWarmWhite}{HTML}{FDFBF7}
\definecolor{RPDarkGrey}{HTML}{171717}
\definecolor{RPLightGreen}{HTML}{628F6B}
\definecolor{RPSkyBlue}{HTML}{8BBCE5}

\newcommand{\lead}[1]{\textbf{#1}}   

\usepackage{duckuments}

\begin{document}

\title[Evaluating Redpine Science]{Leveraging a four-quadrant approach for evaluating Redpine Science}

\author{\fnm{Filip} \sur{Dorm}}

\author{\fnm{Leonora} \sur{Vesterbacka}}

\affil{\orgname{Redpine}, \orgaddress{\country{Sweden}}}

\abstract{Redpine Science gives models and agents a single access point to a wide range of
peer-reviewed literature, queried directly through the Model Context Protocol
(MCP) and an API. This report
evaluates Redpine Science on two levels: the relevance of the retrieved chunks, and a
model's answer when it has access to Redpine Science compared to web search. Both public and expert-validated benchmarks are used.
Public benchmarks are a widely accepted way to test model development and are
comparable across labs, but risk saturation and memorization. To address this, we
complement them with an expert-validated question set. In total, this report
presents four evaluations. On ScholarQABench SciFact, the public answer-quality benchmark
reported here, an agent with Redpine Science answers 94.4\% of claims correctly
against 87.6\% with no retrieval. On the expert-validated question set, an agent with Redpine Science states
80.1\% of the required claims against 70.2\% for an agent restricted to web search. On the 668 queries of a public retrieval benchmark whose gold paper Redpine holds,
stripped of any model reasoning, Redpine Science places the correct source paper in its
top ten results for 83.1\% of queries (Recall@10), against 79.3\% for the benchmark's creator. A blinded expert relevance panel places Redpine Science's
Precision@5 at 75.2\% against 39.8\% for the PubMed search tool. We release the
expert-validated question set and instructions to reproduce every
headline result above, at \url{https://github.com/redpine-ai/benchmarks}.}

\keywords{Evaluation Benchmarks, LLM, RAG, Information Retrieval, Question Answering, Human Evaluation, Biomedical NLP}

\maketitle

\section{Introduction}\label{sec:introduction}

Redpine Science gives models and agents a single access point to a wide range of
peer-reviewed literature, queried directly through the Model Context Protocol
(MCP)~\cite{mcp} and an API. This
report measures how much that access improves the answers a model produces, and how
reliably that improvement can be measured.

Evaluating a retrieval-augmented generation (RAG) system is harder than evaluating a
model alone. A RAG pipeline conflates two effects: whether the right passage is surfaced
(retrieval), and whether the model turns that passage into an accurate, grounded answer
(generation). Most published evaluations report one blended score for both. A win on
that score can come from better retrieval, a better-written answer drawn from similar
evidence, or a mix of the two, and the number alone does not say which.

We separate the two effects by crossing two independent choices: where the evaluation
questions come from, an existing public benchmark or a literature-grounded set built
in-house, and what is measured, the full agentic answer or retrieval alone with no
model reasoning in the loop. This produces four evaluations rather than one, described
in Sections~\ref{sec:method-quadrant-one} through \ref{sec:method-quadrant-four}.
Reporting all four separately shows whether a result comes from retrieval or from
generation.

\subsection{Redpine Science}\label{sec:redpine-science}

The corpus used for these evaluations is Redpine Science, which holds over 20
million peer-reviewed articles from closed-access journals and open-access
repositories, served through Redpine's API and MCP. A query returns a ranked list
of chunks, short passages from inside an article rather than whole documents, each
tied to its source paper.
Coverage concentrates in biomedicine, with smaller coverage of physics and
chemistry. That concentration sets the domain for the rest of this report: the
public benchmarks tested are mostly biomedical, and the relevance judges are PhDs
in a few clinical subspecialties drawn from the corpus's wider biomedical
coverage.\footnote{For the full list of partners and coverage by field, see
\url{https://redpine.ai}.}

\section{Evaluation methodology}\label{sec:methodology}

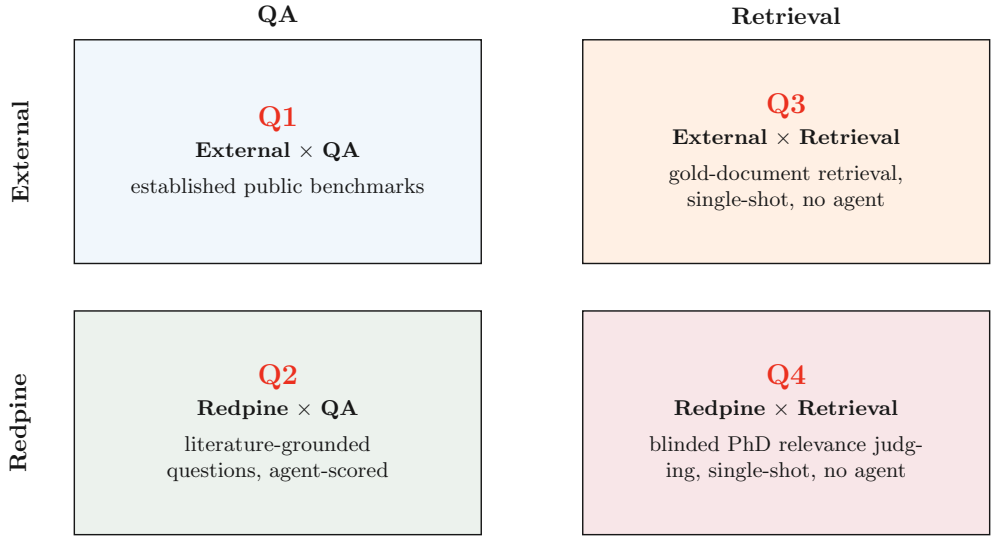
\begin{figure}[h]
\centering
\resizebox{\linewidth}{!}{%
\begin{tikzpicture}[
  quad/.style={draw=RPDarkGrey, line width=0.6pt, text width=5.3cm,
    minimum height=3.3cm, align=center, inner sep=10pt, font=\small, text=RPDarkGrey},
  axis/.style={font=\bfseries, text=RPDarkGrey},
]
  \node[quad, fill=RPSkyBlue!12] (q1) at (0,0)
    {{\bfseries\large\color{RPCrimson}Q1}\\[2pt]
     \textbf{External $\times$ QA}\\[4pt]
     established public benchmarks};
  \node[quad, fill=RPOrange!12] (q3) at (7.5,0)
    {{\bfseries\large\color{RPCrimson}Q3}\\[2pt]
     \textbf{External $\times$ Retrieval}\\[4pt]
     gold-document retrieval, single-shot, no agent};
  \node[quad, fill=RPLightGreen!12] (q2) at (0,-4)
    {{\bfseries\large\color{RPCrimson}Q2}\\[2pt]
     \textbf{Redpine $\times$ QA}\\[4pt]
     literature-grounded questions, agent-scored};
  \node[quad, fill=RPDarkRed!10] (q4) at (7.5,-4)
    {{\bfseries\large\color{RPCrimson}Q4}\\[2pt]
     \textbf{Redpine $\times$ Retrieval}\\[4pt]
     blinded PhD relevance judging, single-shot, no agent};

  \node[axis] at (0,2.0) {QA};
  \node[axis] at (7.5,2.0) {Retrieval};
  \node[axis, rotate=90] at (-3.8,0) {External};
  \node[axis, rotate=90] at (-3.8,-4) {Redpine};
\end{tikzpicture}%
}
\caption{The four evaluations, crossing who built the questions against what is measured.}
\label{fig:methodology}
\end{figure}

The evaluation crosses who built the questions against what is measured, producing the
four evaluations in Figure~\ref{fig:methodology}.

None of these four evaluations is new on its own. Scoring an agent's answer against
public benchmarks or against harder expert-authored questions is standard for RAG and
agent evaluations. For retrieval measured in isolation, the two quadrants follow existing precedent:
information-retrieval datasets that match a query to a gold article, and blinded
expert judging of top retrieved results. Reporting all four together, so that a
retrieval effect and a generation effect are separated for both question sources,
is the contribution of this methodology.

Quadrants one and two score answers from the same kind of agent: a language model in
a tool-use loop, with a fixed budget of tool calls, that decides what to search and
when to stop. Within each of these two quadrants, every arm runs the same loop, the same model, and the same
system prompt; the prompt differs between arms in one clause only, the list of tools
the agent may call. Answers are stored and scored afterwards, so a difference between
arms can come only from the tools. The model, the call budget, the arms and the
scoring rule differ between the two quadrants and are set in each subsection below.

\subsection{External QA benchmark}\label{sec:method-quadrant-one}

Public benchmarks are the standard way to compare models across labs and
domains~\cite{math,swebench,ecollm}: the question set and the scoring rule are
published, so a number can be checked against other groups' results. Biomedicine has many, among them Massive Multitask Language Understanding (MMLU)~\cite{mmlu},
MedHELM~\cite{medhelm}, BioASQ~\cite{bioasq}, MedXpertQA~\cite{medxpertqa},
HealthBench~\cite{healthbench}, LiveMedBench~\cite{livemedbench}, and the Bio/Chem Gold
subset~\cite{hlegold} of Humanity's Last Exam~\cite{hle}.

Many of these benchmarks cannot show a retrieval effect, for two reasons. The first is
saturation. Models answer most of the questions correctly from memory: by 2023, GPT-4
with no retrieval scored 93.8\% on MMLU Professional Medicine and 95.1\% on
MMLU College Biology~\cite{medprompt}. When a model already scores 95\%, retrieval
can add a few points at most, and the run-to-run variation of the model is of the same
size. The second is contamination. The question text has been public for years and is
plausibly in the training data of the models under
test~\cite{sainz2023contamination,deng2024contamination}. A model that has memorized an answer scores well
without retrieving anything, and the benchmark cannot separate the two.

Newer benchmarks reduce both problems: a recent benchmark is less likely to appear in a
model's training data, and one built to be hard for current models leaves room for
retrieval to move the score. ScholarQABench~\cite{openscholar} is a suite of
literature-synthesis and paper-surfacing tasks, use cases Redpine Science serves, so it
is the benchmark reported here. Its SciFact
subtask~\cite{scifact} holds 208 biomedical claims, each a one-sentence scientific
statement such as a drug's effect on a patient group, with a label saying whether the
published literature supports or refutes it. The agent reads the claim, searches if it
has a tool, and states a verdict. The suite's scoring rule marks the answer correct if the
gold label, true or false, appears in the answer text; no judge model is involved.

\lead{Arms.} Claude Sonnet 5 answers every claim in four arms with a budget of 16 tool
calls: no tools, web search, Redpine Science, and Redpine Science with web search as
fallback. In each arm with a tool, the first call is forced to that arm's primary tool,
so no arm answers from memory alone; after that the agent decides. The full set is run
five times and the mean over runs is reported, so the spread is run-to-run variation
rather than one draw.

\subsection{Redpine QA benchmark}\label{sec:method-quadrant-two}

Public benchmarks are comparable across labs, but a model that has memorized an answer
is not retrieving anything. Quadrant two therefore asks questions that are new, that
have one paper behind them, and that a biomedical expert has checked.

\lead{Question set.} The set holds 179 questions in four clinical tracks: cardiology
(79), neuroimmunology (60), sepsis (20) and immunology (20). Each question is written
from one paper in the Redpine Science full-text corpus. Of these papers, 105 are open
access and 74 paywalled. The split matters because web search can find an open-access
paper, so results are reported by access class as well as overall
(Section~\ref{sec:results-quadrant-two}).

\lead{Writing the questions and their answer keys.} A language model pipeline turns a
paper into a question and an answer key. Claude Opus 5 writes and reviews; Claude
Sonnet 5 writes and grades the test answers described below. The pipeline searches
Redpine Science for papers in a track, skipping case reports, protocols without results
and papers with no clinical content. It then writes the question with the paper's
results hidden from it, so that the question reads as one a clinician would ask a
literature assistant rather than as a summary of the paper: one thing asked, 10 to 22
words as the target (three released questions run to 26 after expert review), and no
hint of the paper's design, thresholds or timepoints. The specifics go into
the answer key instead.

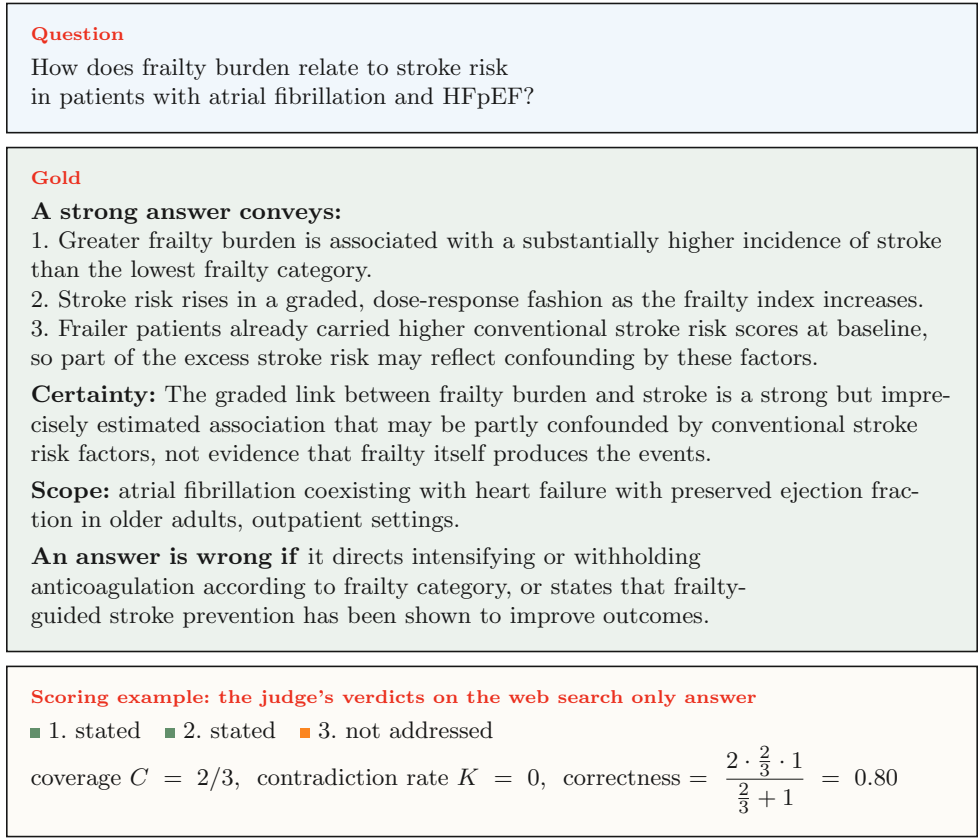
\begin{figure}[htbp]
\centering
\begin{tikzpicture}[
  panel/.style={draw=RPDarkGrey, line width=0.6pt, text width=12.2cm, align=left,
    inner sep=9pt, font=\small, text=RPDarkGrey, anchor=north west},
]
  \newcommand{\paneltag}[1]{{\bfseries\footnotesize\color{RPCrimson}#1}}
  \node[panel, fill=RPSkyBlue!12] (q) at (0,0)
    {\paneltag{Question}\\[2pt]
     How does frailty burden relate to stroke risk in patients with atrial fibrillation
     and HFpEF?};
  \node[panel, fill=RPLightGreen!12] (g) at ([yshift=-5pt]q.south west)
    {\paneltag{Gold}\\[2pt]
     \textbf{A strong answer conveys:}\\
     1.\ Greater frailty burden is associated with a substantially higher incidence of
     stroke than the lowest frailty category.\\
     2.\ Stroke risk rises in a graded, dose-response fashion as the frailty index
     increases.\\
     3.\ Frailer patients already carried higher conventional stroke risk scores at
     baseline, so part of the excess stroke risk may reflect confounding by these
     factors.\\[3pt]
     \textbf{Certainty:} The graded link between frailty burden and stroke is a strong
     but imprecisely estimated association that may be partly confounded by
     conventional stroke risk factors, not evidence that frailty itself produces the
     events.\\[3pt]
     \textbf{Scope:} atrial fibrillation coexisting with heart failure with preserved
     ejection fraction in older adults, outpatient settings.\\[3pt]
     \textbf{An answer is wrong if} it directs intensifying or withholding
     anticoagulation according to frailty category, or states that frailty-guided
     stroke prevention has been shown to improve outcomes.};
  \node[panel, fill=RPWarmWhite] (j) at ([yshift=-5pt]g.south west)
    {\paneltag{Scoring example: the judge's verdicts on the web search only answer}\\[2pt]
     \textcolor{RPLightGreen}{\rule{0.9ex}{0.9ex}}\ 1.\ stated\quad
     \textcolor{RPLightGreen}{\rule{0.9ex}{0.9ex}}\ 2.\ stated\quad
     \textcolor{RPOrange}{\rule{0.9ex}{0.9ex}}\ 3.\ not addressed\\[2pt]
     coverage $C = 2/3$, \ contradiction rate $K = 0$, \ correctness
     $= \dfrac{2 \cdot \tfrac{2}{3} \cdot 1}{\tfrac{2}{3} + 1} = 0.80$};
\end{tikzpicture}
\caption{One question from the cardiology track, its gold, and a worked example of the
scoring: the judge's verdicts on the web search only arm's answer to it, the answer itself
not shown. The gold names three required claims, the certainty and scope
they hold under, and the over-claim an answer must not make. The answer states two of
the three claims and contradicts none, so coverage is 0.67, the contradiction rate is 0
and correctness is 0.80 by Equation~\ref{eq:correctness}. Green marks a stated claim and
orange a claim the answer does not address; a contradicted claim would be marked dark
red.}
\label{fig:q2-example}
\end{figure}

The answer key, which we call the gold, lists two to four required claims, the facts a
good answer must state; Table~\ref{tab:appendix-q2-questions} in the appendix shows four
questions with their golds. Every claim is tied to a quoted passage in the paper, and its
wording is graded to the evidence: a causal claim needs a randomized comparison; anything
weaker is stated as an association or a preliminary finding. The gold also carries a
certainty sentence that says how strong the evidence is, names the population and
setting the claims hold for, and ends with one wrong-if clause where the paper admits
one: the most tempting over-claim, which an answer must not make. Figure~\ref{fig:q2-example} shows one
question with its gold in full. Passages from other papers on the same
question decide whether a finding is corroborated, uncontested or conflicting, which sets
how firmly the gold may state it. A mechanical check confirms that every quote appears in
the paper and every claim has evidence behind it. A model review then judges each gold
as the two specialist reviewers would, calibrated on their verdicts from earlier sets:
each claim is marked supported, overstated, unsupported or contradicted against the
paper's text, and the certainty sentence and wrong-if clause are checked against the
study design. A flagged gold is repaired once and re-validated; one that still fails is
dropped. A question goes to the experts only if this review, a
check that the question belongs to its track, a check that the claims stay faithful to
the paper after repair, and the mechanical check all pass.

The last step measures the gold before an expert sees it. Three test answers are
generated for each question: one that read the paper, one that read a different but
valid source, and one with no retrieval. Each is graded twice for how much of the gold
it covers, a coverage score in percent, and the three scores travel with the question.
A gold on which the paper-informed answer scores below 75\%, or the retrieval-free
answer 50\% or above, is flagged for the expert: the first rewards one paper's phrasing,
the second can be satisfied from memory. The same three answers later guard every edit
to a gold.

\lead{Expert review.} Three PhD-level biomedical experts reviewed the questions in their
own fields, in a browser deck that shows the question, the gold and the evidence behind
it. For each question the expert marked good, edit or bad, could rewrite the question,
and could comment. A suggested rewrite is applied once a check confirms that the gold still
answers it. A change to a gold is accepted only if the three test answers, regraded
against the new gold, keep their coverage scores: the paper-informed answer's score stays
at or above 85\% where it already was and drops by at most 5 points, the other-source answer's score drops by
at most 5 points, the retrieval-free answer's score rises by at most 5 points, and no
test answer
newly contradicts the gold. When either check fails, the
generated version stays and the failed edit is recorded with the reason: a rewrite that
widens the question past what the gold answers, or a gold edit that lets the
retrieval-free answer pass, is refused. A question marked bad is dropped. Of 180 reviewed questions, 125 are unchanged, 40 carry the expert's
wording, 14 carry an edited gold and one was dropped.

\lead{Arms.} Claude Opus 5 answers every question in two arms with a budget of 20
tool calls: web search only, and web search plus Redpine Science. No call is forced;
the agent chooses when to use each tool, which is how a customer's agent would use it.
Every judging pass scores the same stored text.

\lead{Judging.} A second, fixed model (Claude Sonnet 5) judges every answer, the same
way for both arms. Jev, TypeSafe's System One decision model~\cite{jev}, is run as a
second judge over the same answers. It does not generate text:
it takes the question, the gold and the answer as state and returns a typed verdict
with a probability for each claim. Jev's numbers are the mean of two passes; Sonnet 5
is the primary judge, fixed before any expert label was collected. It reads the question, the population and setting from the gold, and
the required claims, and says of each claim whether the answer states it, contradicts
it, or does not address it. Three scores follow. Claim coverage is the share of required
claims the answer states; this is the primary score and corresponds to the strict recall
of TREC nugget evaluation~\cite{nuggetrecall}. Contradiction
rate is the share of claims the answer contradicts, as in K-QA~\cite{kqa}. Correctness combines the two. With $n$ required claims, of which the answer states $s$
and contradicts $c$, coverage is $C = s/n$, the contradiction rate is $K = c/n$, and
\begin{equation}
  \text{correctness} = \frac{2\,C\,(1-K)}{C + (1-K)},
  \label{eq:correctness}
\end{equation}
the harmonic mean of the share of claims stated and the share not contradicted. An
answer that states every claim and contradicts none scores 1; one that states half the
claims and contradicts the rest scores 0.5. Correctness is reported as a percentage
below. The bottom panel of Figure~\ref{fig:q2-example}
works the score through for one real answer: the judge's verdict on each required claim,
and the coverage, contradiction rate and correctness those verdicts give. The judge reads the
answer as written, citations included.

\lead{Checking the judge against the experts.} A study designed before any label was
collected asks whether the judge's claim verdicts agree with the experts'. The same three experts label answers
with the judge's three verdicts plus a fourth label, cannot assess, one answer per screen with its
citations replaced by a placeholder, and the two arms of a question at least two screens
apart. Every expert labels a random core of 22
questions; each also labels six further questions chosen to cover large and small
judge differences, with the sampling weights recorded; eight screens per expert are
repeats for consistency. Agreement is measured with Krippendorff's alpha~\cite{krippendorff} and with the intraclass correlation coefficient (ICC) of the judge
against the experts' consensus, next to the experts' agreement with each other. The
difference between the arms over all 179 questions is estimated by prediction-powered
inference~\cite{ppi,ppiplusplus}, which corrects the judge's
estimate with the expert labels. The sample, the statistics, the reliability floors
(0.80 on whether a claim is stated and 0.667 on the three-way label, Krippendorff's
recommended and tentative standards~\cite{krippendorff2004}) and the decision rules were fixed before the
experts began.

\subsection{External retrieval benchmark}\label{sec:method-quadrant-three}

Quadrants one and two score the full agentic answer, where a poor result
may reflect generation rather than retrieval, and a strong result may reflect writing
quality compensating for weak retrieval. Quadrant three isolates retrieval: each system returns its own ranking once, with no
agent loop, reformulation, or added reasoning.

This quadrant uses the Exa publication-retrieval benchmark~\cite{exabenchmarks}, an
existing open benchmark published by Exa rather than built by us: its queries and gold
documents are public, so the result can be checked. It grounds every query in one
gold publication and scores whether it is surfaced and at what rank: a match is exact
Digital Object Identifier (DOI) when a system returns one or one appears in a result's
URL or text, or a normalized title substring otherwise.

The full benchmark has 1,866 queries; 671 have a gold paper in Redpine's corpus,
selected by corpus membership before any system was run, not by result. Three of these
failed on a transient API error for one system and are excluded rather than scored as
misses, leaving 668, on which all three systems are scored. The excluded queries ask about papers Redpine does not hold, so scoring it on
them would measure corpus coverage rather than retrieval quality, and the numbers here
are not comparable to full-benchmark numbers published elsewhere. The benchmark asks several phrasings per source paper, so these
668 DOI-confirmed queries map to 101 unique source papers. Recall is reported at the
query level, the more demanding number since it counts every phrasing separately, and at
three ranks, Recall@1, Recall@5, and Recall@10, so a gain at the top of the ranking is
distinguishable from one that only appears deeper in it.

Three retrieval services run against it with the identical query and no follow-up
fetch or agent reasoning on any side: Redpine Science, Exa's own search, configured as
Exa's own benchmark implementation runs it~\cite{exabenchmarks}, and Tavily~\cite{tavily}, the web-search arm
used throughout this report. Exa and Tavily, advanced web search services, search the
open web, Redpine Science a curated, licensed corpus, so this compares retrieval
quality on identical terms.

\subsection{Redpine retrieval evaluation}\label{sec:method-quadrant-four}

Quadrant three can only score queries tied to one known publication, but most research
questions have many relevant papers rather than a single gold source. Quadrant four
covers these: a blinded domain expert judges each result for relevance instead of
checking it against a gold paper. Redpine Science is compared against PubMed, a widely used search
engine for the biomedical literature. Both systems receive the identical question text. Redpine Science is
queried once as written; PubMed is queried as written and, when fewer than eight hits
return, re-queried with stopwords and then trailing words removed, keeping the attempt
with the most results. No other reformulation is applied on either side.

Judges score each result on a 0 to 4 scale, 0 not relevant, 4 perfectly relevant,
the scale of a published relevance evaluation~\cite{consensus}, blind to which system produced it; system identity is
randomized left-to-right per query to prevent a judge from guessing rather than
reading. URL, DOI and publication date are removed from every result before a judge
reads it, since a database record and a scraped web page format these differently
enough to identify the source. Two parameters differ from that evaluation: Precision@K
counts a score of 2 or higher as relevant, where it counts 3 or higher, and K is 5 for
both Precision@K and Discounted Cumulative Gain (DCG)@K, where it is 10.

Three PhD-level domain experts are the judges, and the panel holds 115 questions in
two pools. The first pool is 90 subspecialty questions that require domain knowledge
to evaluate the results, not only to answer, in three non-overlapping batches of 30,
each judged by one panelist for breadth across subspecialties: one batch is mostly
neurology, with a handful of adjacent critical care, immunology, microbiology and
pharmacology questions; one is entirely cardiology; one is entirely rheumatology. The
second pool is 25 questions judged by two panelists against the same blinded result
set, to measure inter-rater agreement; per-item scores are averaged across the two
judges before pooling, so each of these questions contributes one data point, not two.

\section{Results}\label{sec:results}

The following subsections present the results of the four evaluations
described above.

\subsection{External QA benchmark}\label{sec:results-quadrant-one}

This quadrant tests whether access to Redpine Science changes an agent's answer
accuracy on a public, claim-verification benchmark.

\begin{figure}[htbp]
    \centering
    \includegraphics[width=0.6\linewidth]{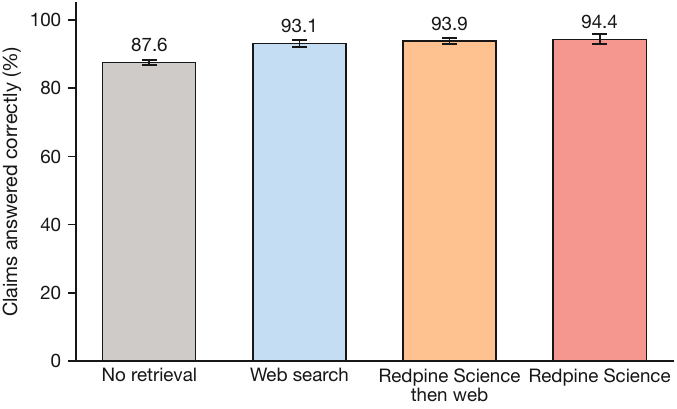}
    \caption{Accuracy on ScholarQABench SciFact by arm, Claude Sonnet 5, mean over
    five runs of the 208 claims. Color encodes the system; error bars are the 95\%
    confidence interval of the mean over the five runs, so they show run-to-run
    variation. A claim is excluded from every arm of the run it falls in when the
    model refuses it with no answer in at least one arm, or when an arm call fails
    after retries, so each run scores 196 to 199 claims.}
    \label{fig:scifact-arms}
\end{figure}

Figure~\ref{fig:scifact-arms}
shows the result. An agent with Redpine Science answers
94.4\% of the claims correctly, against 87.6\% for the same agent with no
retrieval: 6.8 points more, or 7.8\% more claims right, averaged over
the five runs and positive in each of them. Web search scores 93.1\%, and Redpine
Science with web search as fallback 93.9\%.

Every arm scores above 87\%. The no-retrieval score is the floor a retrieval
system starts from, and on this benchmark it leaves 12 points to gain; Redpine Science
closes over half of them. A benchmark on which the model already answers nine claims in ten
from memory can show that retrieval helps, but not by how much, and it says nothing
about whether the retrieved papers are the right ones. The remaining quadrants take
those two questions in turn: quadrant two asks questions the model cannot answer from
memory, and quadrants three and four measure the retrieved papers directly.

\subsection{Redpine QA benchmark}\label{sec:results-quadrant-two}

This quadrant tests answer quality on the expert-validated question set, comparing
an agent with web search alone against the same agent with Redpine Science added.

\begin{figure}[htbp]
    \centering
    \includegraphics[width=\linewidth]{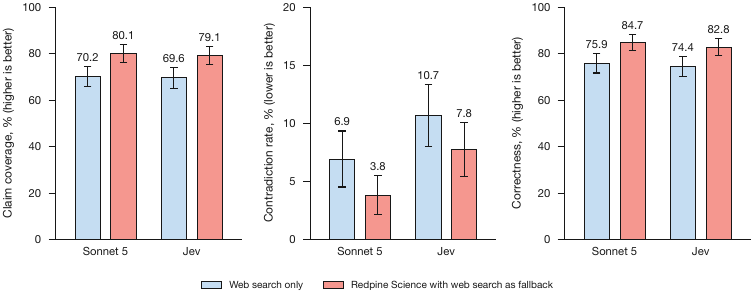}
    \caption{Claim coverage, contradiction rate and correctness by arm on the 179
    expert-validated questions (four tracks), under two judges. Sonnet 5 is the
    primary judge; Jev, averaged over two passes, is a second judge over the
    same answers. Higher is better for coverage and correctness; lower is
    better for the contradiction rate. Error bars are a 95\% confidence interval of the
    mean (Student's $t$) over the 179 questions. The paired intervals for the differences between the arms
    are in the text.}
    \label{fig:quadrant2-metrics}
\end{figure}

\begin{table}[htbp]
\caption{Claim coverage difference in percentage points, web search plus Redpine Science
minus web search only, on all questions and by access class of the source paper,
under each judge. Paired 95\% confidence intervals of the mean (Student's $t$);
n is the number of questions.}
\label{tab:quadrant-two-access}
\begin{tabular}{@{}lrcc@{}}
\toprule
\textbf{Questions} & \textbf{n} & \textbf{Sonnet 5} & \textbf{Jev} \\
\midrule
All questions   & 179 & $+10.0$ $[+5.8, +14.2]$ & $+9.5$ $[+5.6, +13.4]$ \\
Open access     & 105 & $+6.7$ $[+1.7, +11.6]$ & $+5.1$ $[+0.3, +9.9]$ \\
Paywalled       & 74 & $+14.6$ $[+7.2, +22.1]$ & $+15.8$ $[+9.4, +22.1]$ \\
\bottomrule
\end{tabular}
\end{table}

Figure~\ref{fig:quadrant2-metrics} shows the three metrics of
Section~\ref{sec:method-quadrant-two} by arm. Under the primary judge, the arm
with Redpine Science states 80.1\% of the gold claims against 70.2\% for web
search alone, a paired difference of $+10.0$ points (95\% interval $+5.8$ to $+14.2$,
paired t $p = 6\times10^{-6}$, n = 179). It contradicts fewer claims, 3.8\% against
6.9\%, a difference of $-3.1$ points ($-5.5$ to $-0.8$, $p = 0.01$). Correctness,
the harmonic mean of the two per question, averages 84.7\% against 75.9\%, a difference of $+8.8$ points ($+4.9$ to $+12.7$,
$p = 2\times10^{-5}$). Per question, the Redpine Science arm covers more claims on 62
questions, fewer on 22, and the same number on 95. Half the questions are answered
equally well by both arms; the gain sits in the other half.

Jev gives the same picture. Under Jev the coverage difference is
$+9.5$ points ($+5.6$ to $+13.4$, $p = 3\times10^{-6}$), the contradiction difference
$-2.9$ points ($-5.6$ to $-0.2$, $p = 0.03$) and the correctness difference $+8.4$ points
($+4.8$ to $+12.1$, $p = 1\times10^{-5}$), with 62 questions won, 21 lost and 96 tied.
Paired per question, the coverage differences under Jev's first pass and under Sonnet 5
differ by 0.2 points on a standard error of 1.5 points; every comparison of Jev with Sonnet 5
or with the experts below uses that first pass. Jev scores both arms lower on the absolute scale,
mostly because it labels about twice as many claims contradicted (7.8\% against
3.8\% on the Redpine Science arm), but the gap between the arms does not move.
Contradiction is rare under either judge: at most 10.7\% of claims on either arm.
At the claim level the two judges give the same verdict on 89.9\% of claims on the
Redpine Science arm and 87.0\% on the web search arm, with Krippendorff's alpha
of 0.70 and 0.72 on the three-way label. That is agreement between two judges, not
agreement with an expert.

Table~\ref{tab:quadrant-two-access} splits the difference by the access class of the
source paper. On the
74 questions whose paper is paywalled, the Redpine Science arm states 84.9\% of the
gold claims against 70.3\% for web search alone, a difference of $+14.6$ points ($+7.2$ to
$+22.1$, $p = 2\times10^{-4}$, n = 74). On the 105 questions whose paper is open access
the difference is $+6.7$ points ($+1.7$ to $+11.6$, $p = 0.009$, n = 105), from 76.7\%
against 70.1\%. Web search covers the two classes equally; the Redpine Science arm gains
most where web search cannot read the paper. Jev gives the same split, $+15.8$ points on the
paywalled questions and $+5.1$ on the open-access ones. These are the judge's margins;
the expert check follows.

\begin{table}[htbp]
\caption{Expert check on the quadrant two judges. Three domain experts labeled
a blinded sample of the answers with the judge's three verdicts. Judge rows compare
each judge with the experts; panel rows are the experts' agreement with each other.
Majority agreement is over the 144 core claim units on which the expert labels have a
majority; the ten tied units are excluded.
Intervals are 95\%: jackknife for alpha, bootstrap for the intraclass correlation.}
\label{tab:quadrant-two-experts}
\begin{tabular}{@{}lcc@{}}
\toprule
 & \textbf{Sonnet 5} & \textbf{Jev} \\
\midrule
Agreement with the expert majority, 144 claim units & 90\% & 85\% \\
Alt-test winning rate, $\epsilon = 0.2$ & 100\% & 100\% \\
Mean Cohen's $\kappa$ against each expert (expert against expert: 0.48) & 0.59 & 0.55 \\
ICC(2,1) against the leave-one-out consensus & 0.86 [0.77, 0.92] & 0.82 [0.71, 0.90] \\
\midrule
\multicolumn{3}{@{}l}{\textit{Panel, 154 core claim units}} \\
Experts' ICC ceiling & \multicolumn{2}{c}{0.74 [0.56, 0.85]} \\
Krippendorff's alpha, stated versus not & \multicolumn{2}{c}{0.53 [0.42, 0.65]} \\
Krippendorff's alpha, three-way & \multicolumn{2}{c}{0.49 [0.38, 0.59]} \\
Fleiss's $\kappa$, stated versus not / three-way & \multicolumn{2}{c}{0.53 / 0.49} \\
\bottomrule
\end{tabular}
\end{table}

Three domain experts then labeled a blinded sample of the same answers with the
judge's three verdicts, as Section~\ref{sec:method-quadrant-two} describes: a random
core of 22 questions, both arms, every answer labeled by all three experts, and 18 further
questions, both arms, labeled by one expert each. Of the 154 core claim units, 138 carry three
usable labels and 16 carry two, because one expert marked that claim cannot assess. Table~\ref{tab:quadrant-two-experts} reports the agreement.

Both judges agree with the experts at least as well as the experts agree with each other.
Sonnet 5 gives the same verdict as the expert majority on 90\% of the claims with an
expert majority, Jev on 85\%. The alt-test~\cite{alttest} holds out each expert in turn and asks whether the judge agrees with
that expert at least as well as the other two do; at the
$\epsilon = 0.2$ its authors recommend for expert annotators, both judges win for all three. The test gets easier the more a
panel disagrees among itself, so we do not read 100\% as adequacy. Mean pairwise Cohen's $\kappa$
between a judge and an expert is 0.59 for Sonnet 5 and 0.55 for Jev, against 0.48
between two experts. The judges exceed the experts' agreement with each other partly because
the experts disagree among themselves, so this is alignment with a noisy panel, not superiority
to it. On the per-answer coverage score, the intraclass correlation
between Sonnet 5 and the experts' leave-one-out consensus is 0.86, above the experts'
own ceiling of 0.74; Jev's is 0.82. Sonnet 5 is ahead of Jev on every measure by a
small margin.

The expert labels point the same way as the judge's. On the 22-question random core,
where all experts labeled both arms, the expert-only coverage gain is $+24.9$ points
(95\% interval $+11.6$ to $+38.2$); that draw over-represents questions on which the judge saw
a large gap ($+29.9$ against $+10.0$ points on the frame), so the estimate for all 179 questions
is, by prediction-powered
inference with the design's sampling weights, $+15.5$ points ($-1.9$ to
$+33.0$), next to the judge's $+10.0$. Its interval is wide because the design placed only
two labeled questions in the 88-question stratum where the judge saw no gap. The
contradiction rate on the core moves the same way, $-6.3$ points ($-16.1$ to $+3.5$). The
expert labels agree with the judge on the direction of the gain; its size rests on the
judge's 179-question estimate.

The experts' agreement with each other was moderate: Krippendorff's alpha 0.53 on
whether a claim is stated and 0.49 on the three-way label, below the floors of 0.80 and
0.667 we fixed in advance as the condition for calling a judge adequate. No adequacy
verdict is therefore drawn for either judge, and the figures above describe alignment. Disagreement among the experts concentrates on
whether a claim is stated or not addressed, and on the rare contradicted label; it varies
by track, from alpha 0.75 on sepsis and 0.52 on cardiology to 0.23 on neuroimmunology
and 0.20 on immunology (16 claim units).

\subsection{External retrieval benchmark}\label{sec:results-quadrant-three}

This quadrant tests retrieval alone, whether the correct source paper is surfaced,
with no model reasoning involved.

\begin{figure}[h]
  \centering
  \includegraphics[width=0.6\linewidth]{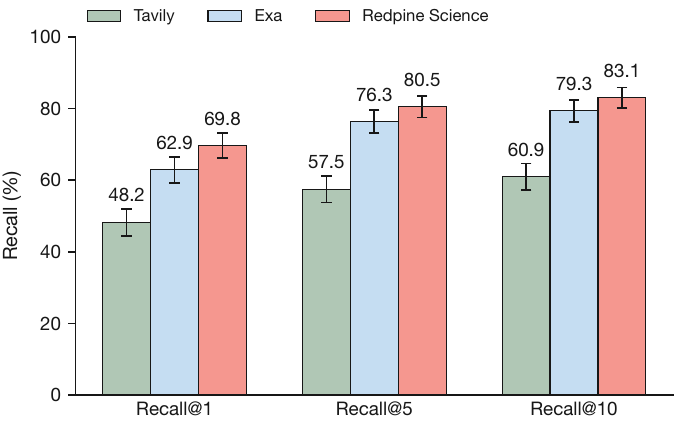}
  \caption{Recall@1, Recall@5, and Recall@10 by system, n = 668 queries whose
  gold paper is confirmed present in Redpine Science's corpus. Error bars are a
  95\% confidence interval for a binomial proportion, computed from each cutoff's
  exact found-count out of 668. Redpine Science leads at every cutoff.}
  \label{fig:quadrant3-recall}
\end{figure}

\begin{table}[h]
\caption{Quadrant three retrieval results, n = 668.}
\label{tab:quadrant-three}
\begin{tabular}{@{}lcccc@{}}
\toprule
 & \textbf{Recall@1} & \textbf{Recall@5} & \textbf{Recall@10} & \textbf{nDCG@10} \\
\midrule
Redpine Science & 69.8\% & 80.5\% & 83.1\% & 0.765 \\
Exa             & 62.9\% & 76.3\% & 79.3\% & 0.714 \\
Tavily          & 48.2\% & 57.5\% & 60.9\% & 0.545 \\
\bottomrule
\end{tabular}
\end{table}

Figure~\ref{fig:quadrant3-recall} and
Table~\ref{tab:quadrant-three} show the result.
Recall@K is the fraction of queries where the correct paper appears in a system's top
K results, on a 0 to 100\% scale. Normalized DCG at ten (nDCG@10) also rewards a higher rank for the correct
paper, on a 0 to 1 scale where 1 means the correct paper is always ranked first.
On the 668 in-corpus queries, Redpine Science leads on all four metrics: 6.9 points
over Exa at Recall@1, narrowing
to 3.7 points at Recall@10, and 21.6 to 23.1 points over Tavily across the three
cutoffs.

\subsection{Redpine retrieval evaluation}\label{sec:results-quadrant-four}

This quadrant tests retrieval relevance directly, judged by a blinded panel of
domain experts rather than by a DOI match.

\begin{figure}[htbp]
    \centering
    \includegraphics[width=0.6\linewidth]{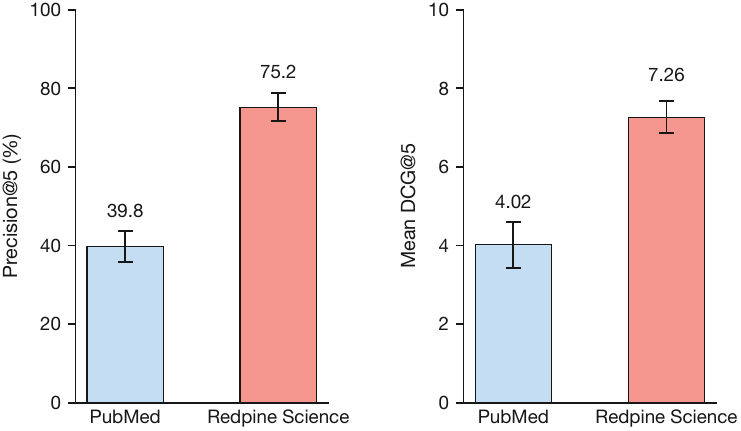}
    \caption{Precision@5 (score 2 or higher on the 0 to 4 relevance scale, at or
    above the midpoint) and mean DCG@5, by system,
    pooled across all 115 questions (90 domain-specific plus 25
    cross-judged). Left panel error bars are a 95\% confidence interval for a
    binomial proportion; right panel error bars are a 95\% confidence interval
    of the mean over the pooled per-query values.}
    \label{fig:quadrant4-metrics}
\end{figure}

Figure~\ref{fig:quadrant4-metrics}
shows the result over all 115 questions.
Precision@5 is the fraction of a system's top 5 results a judge scores 2 or higher out
of 4 for relevance, on a 0 to 100\% scale.
DCG@5 also weights a relevant result more heavily the higher it is ranked; with gains of
0 to 4 its maximum is 11.8, and higher is better. Redpine Science leads on both: Precision@5 75.2\%
against 39.8\% for the PubMed search tool, and mean DCG@5 7.26 against 4.02. Mean relevance per judged
result is 2.43 against 1.38 on the 0 to 4 scale. Agreement between the two panelists
who judged the same set of questions
is moderate (quadratic-weighted Cohen's $\kappa$ = 0.59, n = 246 individually judged
results across the 25 shared questions), the measured inter-rater ceiling for this
panel.

\section{What we are releasing}\label{sec:releasing}

All four evaluations in this report are released as a public benchmark suite, one
directory per evaluation.

\begin{itemize}[leftmargin=1.6em, itemsep=0.5em]
  \item \textbf{External QA benchmark} (\texttt{external-qa}). Its per-arm configuration,
    runner, and the reduced run logs (per-claim scores and exclusions) behind the reported
    numbers.
  \item \textbf{Redpine Biomedical QA} (\texttt{redpine-biomedical-qa}). The 179-question
    expert-validated set with its golds, its configuration and runner, the run log behind
    the reported numbers with every model answer
    (retrieved passage text removed), both judges' code, the per-question judge verdicts
    under both judges, and the agreement between the two judges recomputed from them; the
    experts' claim labels and the second judge's probabilities and confidences are withheld.
  \item \textbf{External retrieval benchmark} (\texttt{external-retrieval}). The
    668-query retrieval dataset, the matching code, and the results behind the reported
    recall.
  \item {\raggedright\textbf{Redpine Biomedical Retrieval Evaluation}
    (\texttt{redpine-biomedical-retrieval}). The 115-question relevance panel's dataset and
    ratings, anonymized, and the scoring code behind the reported precision and DCG.\par}
\end{itemize}

\noindent Both QA benchmarks run on one shared tool-use harness (\texttt{harness}), which
reproduces the logged setup; it is not the internal code that produced the logs.
All four are at \url{https://github.com/redpine-ai/benchmarks}.

\section{Summary}\label{sec:summary}

This report evaluates a retrieval-augmented generation system by separating the
effects of retrieval and generation.

On ScholarQABench SciFact, the agent answers 94.4\% of claims correctly with Redpine
Science against 87.6\% with no retrieval and 93.1\% with web search; every arm scores
above 87\%, so a public benchmark of this kind leaves little room to measure retrieval,
which is what the other three quadrants are for. On the expert-validated question set,
an agent with Redpine Science states 80.1\% of the required claims against 70.2\% for
an agent restricted to web search, a paired gain of 10.0 points in claim coverage under
the primary judge; expert labels on a blinded sample point the same way. On the 668
queries of the public retrieval benchmark whose gold paper Redpine holds, Redpine
Science places the correct document in its top ten results for 83.1\% of queries
(Recall@10), against 79.3\% for Exa, the benchmark's creator. On the expert-judged
retrieval panel, three PhD-level domain experts (neurology, cardiology, and
rheumatology) plus a second, cross-judged pool of questions, n = 115 in all, place
Redpine Science's Precision@5 at 75.2\% against 39.8\% for the PubMed search tool.

\section{Acknowledgments}\label{sec:thanks}

We thank the creators of the many public benchmarks this report draws on and
compares against; every comparison in this report against other labs' results
is possible because of their decision to build and publish that work openly.
Evaluation infrastructure of this kind is a shared resource.

We thank our expert judges for their time and care, and Felix Martin for early
discussions that laid the foundation for this evaluation work.

\begin{appendices}

\section{Example Questions}\label{sec:appendix-questions}

Table~\ref{tab:appendix-q2-questions} shows four of the 179 expert-validated questions
of quadrant two (Section~\ref{sec:method-quadrant-two}), one per track, each with the
required claims and the scope line of its gold, quoted as the gold states them. The full
set is released. Table~\ref{tab:appendix-questions} shows one example question from each
of the three domains in quadrant four's judging panel
(Section~\ref{sec:method-quadrant-four}).

\begin{table}[h]
\caption{One example question per domain, quadrant four's judging panel.}
\label{tab:appendix-questions}
\begin{tabularx}{\textwidth}{@{}llX@{}}
\toprule
\textbf{Domain} & \textbf{Subspecialty} & \textbf{Question} \\
\midrule
Neurology & Motor neuron disease & Does prior military service increase the risk
of developing ALS? \\[0.6em]
Cardiology & Cardiac imaging & In patients with STEMI undergoing primary PCI, is
CMR-measured infarct size reliable for predicting 1-year heart failure
hospitalization and mortality? \\[0.6em]
Rheumatology & RA immunology and pathogenesis & How does methotrexate treatment
affect peripheral blood B cells, such as the B cell repertoire and the frequency
of B cell phenotype subtypes? \\
\bottomrule
\end{tabularx}
\end{table}

\begin{table}[h]
\caption{Four of the 179 quadrant two questions, one per track, with the required claims
and scope of each gold. The full set is released.}
\label{tab:appendix-q2-questions}
\begin{tabularx}{\textwidth}{@{}lXX@{}}
\toprule
\textbf{Track} & \textbf{Question} & \textbf{Required claims and scope} \\
\midrule
Cardiology & How does in-hospital mortality compare between transcatheter and surgical
aortic valve replacement in patients with aortic stenosis? & In-hospital mortality was
lower after transcatheter than after surgical aortic valve replacement overall; among
patients younger than 75 years, in-hospital mortality did not differ between the
transcatheter and surgical approaches. Scope: applies to severe non-rheumatic aortic
stenosis treated with aortic valve replacement in adults, predominantly elderly,
inpatient settings. \\[0.6em]
Neuroimmunology & Which imaging measures relate to disability progression that occurs
independently of relapses in multiple sclerosis? & Measures of cortical damage are
central to predicting progression of disability that occurs without relapses; spinal
cord degeneration also carries prognostic relevance for disability progression in
multiple sclerosis; cortical thickness in the temporal lobe emerged as the strongest
imaging predictor of time to relapse-independent progression. Scope: applies to multiple
sclerosis in adults under regular clinical evaluation, outpatient settings. \\[0.6em]
Sepsis & How does the timing of norepinephrine initiation after hypotension onset relate
to mortality in sepsis? & A longer delay between hypotension onset and the start of
norepinephrine is associated with higher 28-day mortality; the delay remains associated
with 28-day death after multivariable adjustment for measured covariates; delays
extending beyond the first hour after hypotension onset are associated with a higher
adjusted hazard of death by day 28. Scope: applies to sepsis with hypotension requiring
vasopressor support in adults, mixed settings. \\[0.6em]
Immunology & How do ibrutinib, acalabrutinib, and zanubrutinib compare for atrial
fibrillation risk in mantle cell lymphoma? & Newly recorded atrial fibrillation or
flutter is less frequent with acalabrutinib than with ibrutinib, on the order of a third
as frequent; newly recorded atrial fibrillation or flutter is likewise less frequent with
zanubrutinib than with ibrutinib; atrial fibrillation or flutter frequency is similar
between acalabrutinib and zanubrutinib, with no clear separation between the two
second-generation agents. Scope: applies to mantle cell lymphoma treated with a covalent
Bruton tyrosine kinase inhibitor in adults, outpatient settings. \\
\bottomrule
\end{tabularx}
\end{table}

\end{appendices}

\clearpage
\bibliography{references}

\end{document}